\documentclass[smallextended]{svjour3} 
\smartqed  
\usepackage[utf8]{inputenc}
\usepackage{algorithm}
\usepackage[noend]{algpseudocode}

\algrenewcommand{\algorithmiccomment}[1]{\hskip3px$\#$ #1}
\algdef{SE}[SUBALG]{Indent}{EndIndent}{}{\algorithmicend\ }%
\algtext*{Indent}
\algtext*{EndIndent}
\usepackage{url}
\usepackage[affil-sl]{authblk}
\usepackage{graphicx}

\begin{document}

\title{Symbolic-Regression Boosting}
\author{Moshe Sipper \and Jason H. Moore}

\institute{M. Sipper \at
              Department of Computer Science, Ben-Gurion University, Beer Sheva 84105, Israel \\
              \email{sipper@gmail.com}           
           \and
           J. H. Moore \at
              Institute for Biomedical Informatics, University of Pennsylvania, Philadelphia, PA 19104-6021, USA \\
        This is a preprint of an article published in Genetic Programming and Evolvable Machines. The final authenticated version is available online at  \texttt{https://link.springer.com/10.1007/s10710-021-09400-0}
}

\date{Received: \today / Accepted: date}





\maketitle

\begin{abstract}
Modifying standard gradient boosting by replacing the embedded weak learner in favor of a strong(er) one, we present SyRBo: Symbolic-Regression Boosting. Experiments over 98 regression datasets show that by adding a small number of boosting stages---between 2--5---to a symbolic regressor, statistically significant improvements can often be attained. We note that coding SyRBo on top of any symbolic regressor is straightforward, and the added cost is simply a few more evolutionary rounds. SyRBo is essentially a simple add-on that can be readily added to an extant symbolic regressor, often with beneficial results.
\end{abstract}

\section{Introduction}
\label{sec-intro}
In machine learning, a \textit{weak learner} is defined as a learner that can produce an hypothesis that performs only slightly better than random guessing, while a \textit{strong learner} can with high probability output an hypothesis that is correct on all but an arbitrarily small fraction of the instances. 

In his seminal paper, ``The strength of weak learnability'', Schapire \cite{schapire1990strength} described a method ``for converting a weak learning algorithm into one that achieves arbitrarily high accuracy.''
Over the years, a plethora of highly successful \textit{boosting} algorithms that transform weak learners into strong ones have been devised \cite{Chen:2016,freund1997decision,friedman2001greedy,LightGBM2017}.

A recent rigorous benchmarking study of four symbolic regression algorithms versus nine machine learning approaches found that ``symbolic regression performs strongly compared to state-of-the-art gradient boosting algorithms'' (they also found that ``in terms of running times [symbolic regression] is among the slowest of the available methodologies'') \cite{orzechowski2018}. 
Herein we wish to combine boosting with symbolic regression, asking whether gradient boosting might improve a strong(er) learner in the form of a symbolic regressor. We answer in the affirmative, demonstrating that improved results can be readily obtained, at relatively little added cost.

In the next section we describe our method and discuss related work. Section~\ref{sec-exp} presents the experimental setup and our results, followed by concluding remarks in Section~\ref{sec-conc}.

\section{Symbolic-Regression Boosting (SyRBo)}
\label{sec-syrbo}
For our experiments we used the popular scikit-learn Python package \cite{scikit-learn,sklearn-website} due to its superb ability to handle much of the tedious desiderata of machine learning coding and experimentation. We then chose the GPLearn package  \cite{stephens2019gplearn}, which implements tree-based genetic programming (GP) symbolic regression, is relatively fast, and---importantly---interfaces seamlessly with scikit-learn. 

The main idea behind SyRBo is simple: we replace the boosted weak learner of gradient boosting (typically a decision tree) with a (possibly) strong learner, specifically, a GP-based symbolic regressor. 

Algorithm~\ref{alg:syrbo} provides the pseudocode  (the code is available at \url{https://github.com/moshesipper}). SyRBo receives the number of boosting stages as a parameter (one might consider the actual number of stages to be one less, as the first stage performs the initial prediction). Fitting a model to data is done in a standard gradient-boosting manner, through successive stages, where each stage fits a learner to the pseudo-residuals of the previous stage; prediction is performed by summing up all learner predictions.  The only change involves the learners themselves, which are not decision trees but rather symbolic regressors, evolved by calling the \textit{SymbolicRegressor} function with the given population size and generation count (both set to 200). The function set used by \textit{SymbolicRegressor} is given in Table~\ref{tab:fset}.
To facilitate the symbolic regressor’s handling of diverse features scales, the dataset rows undergo L2 normalization (i.e., the feature values in a row have a unit L2
norm).

\begin{algorithm}
\caption{SyRBo.}\label{alg:syrbo}
\begin{algorithmic}[1]
\Statex
\Require
\Indent
\Statex \textit{stages} $\gets$ number of boosting stages
\Statex \textit{population\_size} $\gets$ 200
\Statex \textit{generations} $\gets$ 200
\EndIndent

\Statex

\Function{init}{\textit{stages}, \textit{population\_size}, \textit{generations}}
    \State Initialize an empty SyRBo object with given parameters
    \State \textit{boosters} = \{\} \Comment{Initialize an empty list of boosters}
\EndFunction
\Statex
\Function{fit}{\textit{X}, \textit{y}} \Comment{\textit{X}: training inputs, \textit{y}: target values}
    \For{ \textit{stage} $\gets$ 0 to \textit{stages-1} } 
        \State \textit{gp} = \textit{SymbolicRegressor(population\_size, generations)} \Comment{Initialize a GP regressor}
        \State \textit{gp.fit(X,y)} \Comment{Fit regressor to (training) data}
        \State Add \textit{gp} to \textit{boosters} \Comment{Add the fitted GP regressor to the list of boosters}
        \State \textit{y} = \textit{y} - \textit{gp.predict(X)} \Comment{Compute pseudo-residuals}
    \EndFor
\EndFunction
\Statex
\Function{predict}{\textit{X}} \Comment{\textit{X}: inputs}
    \State \textit{prediction} = \textbf{0} \Comment{Vector of zeros whose length equals number of instances in dataset}
    \For{ \textit{i} $\gets$ 0 to \textit{stages-1} } 
        \State \textit{prediction} = \textit{prediction} + \textit{\textit{boosters[i].predict(X)}}
    \EndFor
    \State Return \textit{prediction} 
\EndFunction

\end{algorithmic}
\end{algorithm} 

\begin{table}
\caption{Function set used by \textit{SymbolicRegressor}.}
\label{tab:fset}         
\centering
\vspace{5pt}
\resizebox{0.95\textwidth}{!}{%
\begin{tabular}{r|c|l}
\textbf{Function} & \textbf{Arity} & \textbf{Description} \\ \hline 
add  & 2 & addition \\
sub  & 2 & subtraction \\
mul  & 2 & multiplication \\
div  & 2 & protected division (near-zero denominator returns 1) \\
sqrt & 1 & protected square root (uses absolute value of argument)  \\
log  & 1 & protected log (uses absolute value of argument, near-zero argument returns 0) \\
abs  & 1 & absolute value \\
neg  & 1 & negative  \\
inv  & 1 & protected inverse (near-zero argument returns 0) \\
max  & 2 & maximum \\
min  & 2 & minimum \\
if3  & 3 & $if3(x1,x2,x3)$ returns $x2$ if $x1 \geq 0$ else returns $x3$ \\
if4  & 4 & $if4(x1,x2,x3,x4)$ returns $x3$ if $x1 \geq x2$ else returns $x4$ \\
\end{tabular}
}
\end{table}

We note that our aim herein was to demonstrate SyRBo’s being an add-on that can be added to any symbolic regressor. As such, this paper is not about symbolic regression per se, but about performance benefits to be gained if one is using it. We thus contented ourselves with the standard function set of gplearn (adding only 2 conditionals), with all other parameters set to defaults (except for population size and generation count).

Regarding previous work, it would seem that by and large the emphasis in boosting techniques has been on weak learners, typically decision trees. Works using strong learners in the context of boosting employed mainly AdaBoost-like \cite{freund1997decision} boosting. Modest success was attained by \cite{fink2004mutual,Harries1999}. \cite{wickramaratna2001performance} showed that boosting a strong learner with AdaBoost may, in fact, contribute to performance degradation. Within the domain of GP, AdaBoost-like boosting of dataset sample weights has been used with some success \cite{iba1999bagging,karakativc2018building,oliveira2006using,paris2001applying}.
Perhaps closest to our work is that of \cite{Oliveira2015}, who presented an interesting iterative approach, Sequential Symbolic Regression, wherein each iteration applies a transformation based on a geometric semantic crossover operator. In contrast, our work is based on gradient boosting, is more generic in that it can work with any form of symbolic regression, and is also easier to code and apply to any extant project.

\section{Experimental Setup and Results}
\label{sec-exp}

\begin{figure}
\centering
\includegraphics[scale=0.8]{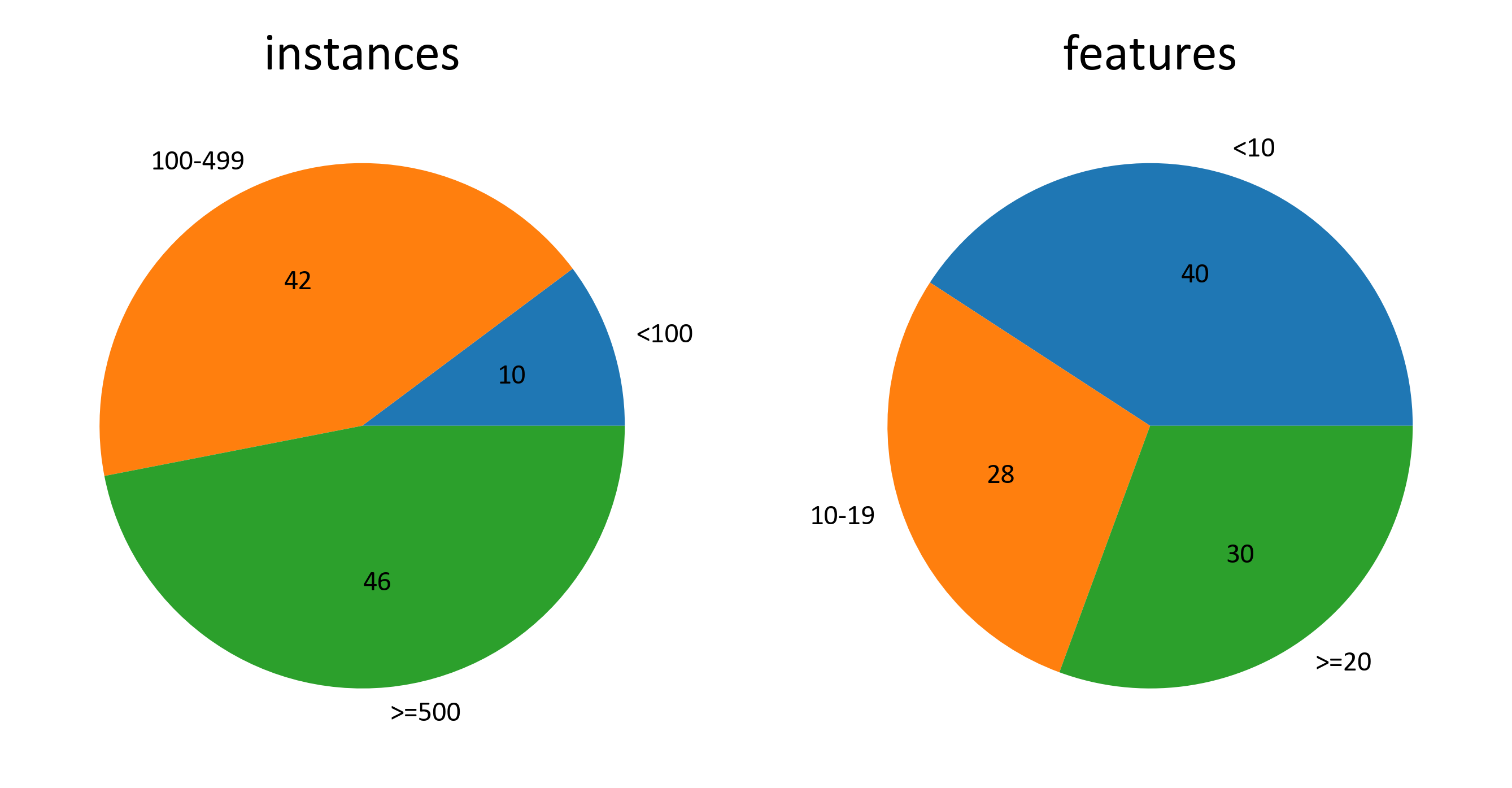}
\caption{A ``bird's-eye view'' of the 98 datasets used in this study: number of instances (left) and number of features (right).}
\label{fig:ds-pie}
\end{figure}

Can this (fairly) simple gradient boosting-like setup improve symbolic regression?
We tested SyRBo on regression datasets from the PMLB repository \cite{orzechowski2018}, using our cluster of Intel\textsuperscript{\textregistered} Xeon\textsuperscript{\textregistered} E5-2650L servers. Of the 120 datasets we selected the 98 with 3000 instances or less.
Figure~\ref{fig:ds-pie} shows a ``bird's-eye view'' of the datasets.

The pseudo-code for the experimental setup is given in Algorithm~\ref{alg:setup}. 
For each dataset we performed 30 replicate runs, with 5-fold cross validation per replicate. 
SyRBo and SymbolicRegressor (with equal population size and generations) were trained on 4 folds and tested on the left-out test fold.

\begin{algorithm}
\caption{Experimental setup.}\label{alg:setup}
\begin{algorithmic}[1]
\Statex
\Require
\Indent
\Statex \textit{dataset} $\gets$ dataset to be used
\Statex \textit{algorithms} $\gets$ \{SyRBo, SymbolicRegressor\}
\EndIndent

\Ensure
\Indent
\Statex Performance measures (over test sets)
\EndIndent

\Statex

\For{ \textit{rep} $\gets$ 1 to \textit{30} } 
    \State Shuffle \textit{dataset} and generate 5 folds
    \For{ \textit{fold} $\gets$ 1 to \textit{5} } 
        \State Split \textit{dataset} into \textit{training} and \textit{test} sets according to \textit{fold}
        \For{ \textit{alg} in \textit{algorithms} }
            \State Use \textit{alg} to fit a \textit{model} to \textit{training} set 
            \State Test resultant \textit{model} on \textit{test} set
        \EndFor
    \EndFor
\EndFor
\end{algorithmic}
\end{algorithm} 

For each of the 98 datasets we recorded the mean absolute error attained per algorithm over each of the 30 replicate runs, per each of the 5 test folds (i.e., following training). 
We ran 4 separate experiments, over all 98 datasets, with number of stages equal to 2, 3, 4, and 5, respectively.

Table~\ref{tab:results} shows a summary of our results (detailed results can be found in the Appendix).
For each dataset we computed the median of the test scores of all 30 replicates, with 5 folds per replicate (a total of 150 test-score values). 
A win for SyRBo was then a better (lower) median value than SymbolicRegressor.
To assess whether a win for a specific dataset was significant or not, we performed a 10,000-round permutation test, comparing the scores of SyRBo with SymbolicRegressor; if the p-value was $< 0.05$ the win was considered significant, else it was not (in which case SyRBo was at least not performing worse than SymbolicRegressor).
In addition, when SyRBo ``lost'' to SymbolicRegressor we performed a 10,000-round permutation test, comparing the two algorithms'  scores; if the p-value was $>= 0.05$ the loss was considered insignificant.

\begin{table}
    \caption{Results. Datasets: number of datasets. Stages: number of stages. Wins: number of datasets for which SyRBo's result was better than SymbolicRegressor's. Significant: number of datasets for which SyRBo's win was significant according to permutation testing. Losses: number of datasets for which SyRBo's result was worse than SymbolicRegressor's. Insignificant: number of datasets for which SyRBo's loss was insignificant according to permutation testing. }
    \centering
    \begin{tabular}{c|c|c|c|c|c}
        Datasets &  Stages & Wins & Significant & Losses & Insignificant \\ \hline
           98 & 2 & 78 & 48 & 20 & 16 \\                
           98 & 3 & 83 & 63 & 15 & 13 \\                
           98 & 4 & 84 & 71 & 14 & 12 \\                
           98 & 5 & 87 & 70 & 11 & 9  \\                
    \end{tabular}
    \label{tab:results}
\end{table}

As seen in the table, statistically significant improvements can often be attained, and, moreover, rarely does SyRBo result in statistically significant worse results. Using SyRBo is thus a good bet, and, furthermore, it is easily coded and the added computational cost is not high.

\section{Concluding Remarks}
\label{sec-conc}
We presented SyRBo, a gradient boosting-style algorithm, wherein the decision tree is replaced by a symbolic regressor. Testing the merits of our new method we showed that symbolic regression results can be consistently improved.
Moreover, as can be seen in Algorithm~\ref{alg:syrbo}, coding SyRBo on top of any symbolic regressor is straightforward, and the added cost is simply a few more evolutionary rounds. SyRBo is essentially a simple add-on that can be readily added to a symbolic regressor, often with beneficial results.

There are a number of avenues we can offer for future exploration:
\begin{itemize}
    \item Add known boosting tricks of the trade, such as a learning rate and dynamic early stopping (similar to XGBRegressor's `early stopping rounds' parameter).
    
    \item Our focus herein was on regression. It would seem worthwhile to examine SyRBo for classification. 
    
    \item We used a rather basic symbolic regressor. The GP literature is rife with many other types of regressors, which might be used in conjunction with SyRBo. More generally, other types of GP might offer productive ways to evolve programs that might serve as strong learners.
    
    \item Comparison to non-symbolic-regressor methods.
    
    \item While we focused on gradient boosting, other types of boosting techniques might be examined as to whether they might be a good fit for SyRBo.
\end{itemize}

\section*{Acknowledgments}
This work was supported by National Institutes of Health (USA) grants LM010098, LM012601, AI116794.
We thank Hagai Ravid for spotting an error in an earlier version of the code.

\bibliographystyle{spmpsci}
\bibliography{references}

\section*{Appendix: Detailed Results}

The results of all experiments over all datasets are given in Tables~\ref{tab:allres2}, ~\ref{tab:allres3}, ~\ref{tab:allres4}, and~\ref{tab:allres5} for number of stages equal to 2, 3, 4, and 5, respectively.
As noted in Section~\ref{sec-exp}, for each of the 98 datasets we recorded the mean absolute error attained per algorithm over each of the 30 replicate runs, per each of the 5 test folds. We then computed the median of these scores, which are presented under `mean absolute error' in the tables. Under `pval' we show the results of the 10,000-round permutation tests between the scores of SyRBo and SymbolicRegressor, with a `!' denoting a significant win for SyRBo and a `=' denoting an insignificant loss for SyRBo.
Under `run times' we show the median run times for SyRBo and SymbolicRegressor. `SR' denotes SymbolicRegressor.

\begin{table}[p]
\caption{2-stage SyRBo: Results of all datasets.}
\label{tab:allres2}
\centering
\tiny
\begin{tabular}{r|c|l}
  \textbf{dataset} & \textbf{mean absolute error and pval} & \textbf{run times} \\ \hline
1027\_ESL & SyRBo: 1.02, SR: 1.04, pval: 9.5E-02 & SyRBo: 59.89s, SR: 31.17s \\
1028\_SWD & SyRBo: 0.61, SR: 0.62, pval: 2.2E-01 & SyRBo: 46.61s, SR: 23.87s \\
1029\_LEV & SyRBo: 0.62, SR: 0.65, pval: 0.0E+00 ! & SyRBo: 54.6s, SR: 26.46s \\
1030\_ERA & SyRBo: 1.45, SR: 1.46, pval: 3.9E-01 & SyRBo: 61.8s, SR: 30.06s \\
1089\_USCrime & SyRBo: 25.59, SR: 27.52, pval: 2.3E-01 & SyRBo: 123.68s, SR: 73.81s \\
1096\_FacultySalaries & SyRBo: 3.56, SR: 3.63, pval: 6.2E-01 & SyRBo: 61.05s, SR: 36.83s \\
192\_vineyard & SR: 2.44, SyRBo: 2.52, pval: 2.1E-01 = & SyRBo: 66.24s, SR: 37.82s \\
195\_auto\_price & SR: 1990.78, SyRBo: 2090.58, pval: 8.9E-02 = & SyRBo: 316.35s, SR: 195.05s \\
207\_autoPrice & SR: 1955.88, SyRBo: 2093.54, pval: 2.8E-02 & SyRBo: 424.44s, SR: 176.55s \\
210\_cloud & SR: 0.51, SyRBo: 0.52, pval: 4.1E-01 = & SyRBo: 44.47s, SR: 22.64s \\
228\_elusage & SyRBo: 12.97, SR: 13.72, pval: 3.2E-01 & SyRBo: 126.28s, SR: 89.94s \\
229\_pwLinear & SyRBo: 1.56, SR: 1.66, pval: 1.2E-02 ! & SyRBo: 65.29s, SR: 35.59s \\
230\_machine\_cpu & SyRBo: 43.04, SR: 45.95, pval: 1.2E-01 & SyRBo: 166.36s, SR: 92.74s \\
4544\_GeographicalOriginalofMusic & SR: 0.49, SyRBo: 0.5, pval: 8.2E-02 = & SyRBo: 62.38s, SR: 36.86s \\
485\_analcatdata\_vehicle & SyRBo: 151.48, SR: 179.22, pval: 1.5E-03 ! & SyRBo: 220.1s, SR: 135.99s \\
505\_tecator & SR: 5.17, SyRBo: 5.83, pval: 1.0E-02 & SyRBo: 92.7s, SR: 51.74s \\
519\_vinnie & SR: 1.27, SyRBo: 1.31, pval: 1.3E-02 & SyRBo: 64.34s, SR: 35.1s \\
522\_pm10 & SyRBo: 0.68, SR: 0.69, pval: 2.1E-01 & SyRBo: 44.3s, SR: 22.12s \\
523\_analcatdata\_neavote & SR: 0.51, SyRBo: 0.52, pval: 7.6E-01 = & SyRBo: 59.63s, SR: 35.95s \\
527\_analcatdata\_election2000 & SR: 38724.65, SyRBo: 39617.02, pval: 7.8E-01 = & SyRBo: 562.68s, SR: 230.48s \\
542\_pollution & SR: 176.33, SyRBo: 185.12, pval: 5.9E-01 = & SyRBo: 215.87s, SR: 141.38s \\
547\_no2 & SR: 0.58, SyRBo: 0.58, pval: 8.7E-01 = & SyRBo: 57.51s, SR: 29.23s \\
556\_analcatdata\_apnea2 & SR: 825.75, SyRBo: 840.89, pval: 7.3E-01 = & SyRBo: 164.27s, SR: 68.58s \\
557\_analcatdata\_apnea1 & SyRBo: 828.37, SR: 844.69, pval: 7.1E-01 & SyRBo: 130.15s, SR: 50.06s \\
560\_bodyfat & SR: 4.21, SyRBo: 4.64, pval: 4.0E-04 & SyRBo: 59.26s, SR: 33.41s \\
561\_cpu & SyRBo: 30.99, SR: 34.95, pval: 8.4E-02 & SyRBo: 115.22s, SR: 77.02s \\
579\_fri\_c0\_250\_5 & SyRBo: 0.42, SR: 0.45, pval: 0.0E+00 ! & SyRBo: 53.18s, SR: 26.01s \\
581\_fri\_c3\_500\_25 & SyRBo: 0.71, SR: 0.72, pval: 2.3E-03 ! & SyRBo: 43.32s, SR: 21.5s \\
582\_fri\_c1\_500\_25 & SyRBo: 0.7, SR: 0.72, pval: 3.8E-02 ! & SyRBo: 53.82s, SR: 26.82s \\
583\_fri\_c1\_1000\_50 & SyRBo: 0.73, SR: 0.74, pval: 1.5E-02 ! & SyRBo: 61.81s, SR: 30.2s \\
584\_fri\_c4\_500\_25 & SyRBo: 0.7, SR: 0.71, pval: 3.6E-02 ! & SyRBo: 53.68s, SR: 26.69s \\
586\_fri\_c3\_1000\_25 & SyRBo: 0.69, SR: 0.71, pval: 6.7E-03 ! & SyRBo: 54.44s, SR: 27.13s \\
588\_fri\_c4\_1000\_100 & SyRBo: 0.73, SR: 0.73, pval: 8.7E-01 & SyRBo: 41.83s, SR: 21.16s \\
589\_fri\_c2\_1000\_25 & SyRBo: 0.7, SR: 0.71, pval: 2.2E-02 ! & SyRBo: 53.3s, SR: 26.58s \\
590\_fri\_c0\_1000\_50 & SyRBo: 0.39, SR: 0.4, pval: 1.0E-02 ! & SyRBo: 55.05s, SR: 28.77s \\
591\_fri\_c1\_100\_10 & SyRBo: 0.71, SR: 0.73, pval: 1.6E-01 & SyRBo: 55.02s, SR: 26.14s \\
592\_fri\_c4\_1000\_25 & SyRBo: 0.72, SR: 0.72, pval: 2.2E-01 & SyRBo: 52.62s, SR: 26.27s \\
593\_fri\_c1\_1000\_10 & SyRBo: 0.65, SR: 0.71, pval: 0.0E+00 ! & SyRBo: 54.59s, SR: 26.25s \\
594\_fri\_c2\_100\_5 & SyRBo: 0.64, SR: 0.68, pval: 1.9E-02 ! & SyRBo: 55.12s, SR: 27.14s \\
595\_fri\_c0\_1000\_10 & SyRBo: 0.39, SR: 0.44, pval: 0.0E+00 ! & SyRBo: 54.06s, SR: 27.02s \\
596\_fri\_c2\_250\_5 & SyRBo: 0.63, SR: 0.68, pval: 0.0E+00 ! & SyRBo: 45.99s, SR: 22.61s \\
597\_fri\_c2\_500\_5 & SyRBo: 0.61, SR: 0.67, pval: 0.0E+00 ! & SyRBo: 55.13s, SR: 26.54s \\
598\_fri\_c0\_1000\_25 & SyRBo: 0.41, SR: 0.43, pval: 2.0E-04 ! & SyRBo: 53.61s, SR: 27.51s \\
599\_fri\_c2\_1000\_5 & SyRBo: 0.57, SR: 0.67, pval: 0.0E+00 ! & SyRBo: 55.04s, SR: 26.24s \\
601\_fri\_c1\_250\_5 & SyRBo: 0.58, SR: 0.65, pval: 0.0E+00 ! & SyRBo: 55.37s, SR: 26.62s \\
602\_fri\_c3\_250\_10 & SyRBo: 0.69, SR: 0.72, pval: 7.3E-03 ! & SyRBo: 52.95s, SR: 25.94s \\
603\_fri\_c0\_250\_50 & SyRBo: 0.4, SR: 0.4, pval: 9.6E-01 & SyRBo: 42.83s, SR: 22.17s \\
604\_fri\_c4\_500\_10 & SyRBo: 0.68, SR: 0.72, pval: 0.0E+00 ! & SyRBo: 43.33s, SR: 21.33s \\
605\_fri\_c2\_250\_25 & SyRBo: 0.69, SR: 0.69, pval: 7.7E-01 & SyRBo: 44.26s, SR: 21.92s \\
606\_fri\_c2\_1000\_10 & SyRBo: 0.64, SR: 0.67, pval: 0.0E+00 ! & SyRBo: 54.98s, SR: 26.73s \\
607\_fri\_c4\_1000\_50 & SyRBo: 0.72, SR: 0.73, pval: 1.7E-01 & SyRBo: 43.72s, SR: 21.98s \\
608\_fri\_c3\_1000\_10 & SyRBo: 0.66, SR: 0.7, pval: 0.0E+00 ! & SyRBo: 57.53s, SR: 28.12s \\
609\_fri\_c0\_1000\_5 & SyRBo: 0.41, SR: 0.44, pval: 0.0E+00 ! & SyRBo: 42.59s, SR: 20.94s \\
611\_fri\_c3\_100\_5 & SyRBo: 0.64, SR: 0.66, pval: 3.1E-01 & SyRBo: 56.53s, SR: 27.55s \\
612\_fri\_c1\_1000\_5 & SyRBo: 0.59, SR: 0.69, pval: 0.0E+00 ! & SyRBo: 56.08s, SR: 26.31s \\
613\_fri\_c3\_250\_5 & SyRBo: 0.59, SR: 0.62, pval: 0.0E+00 ! & SyRBo: 54.03s, SR: 26.4s \\
615\_fri\_c4\_250\_10 & SyRBo: 0.67, SR: 0.7, pval: 5.6E-03 ! & SyRBo: 51.54s, SR: 25.62s \\
616\_fri\_c4\_500\_50 & SyRBo: 0.74, SR: 0.74, pval: 9.6E-01 & SyRBo: 52.4s, SR: 26.34s \\
617\_fri\_c3\_500\_5 & SyRBo: 0.58, SR: 0.65, pval: 0.0E+00 ! & SyRBo: 54.41s, SR: 26.8s \\
618\_fri\_c3\_1000\_50 & SyRBo: 0.72, SR: 0.73, pval: 1.0E-01 & SyRBo: 51.91s, SR: 26.01s \\
620\_fri\_c1\_1000\_25 & SyRBo: 0.72, SR: 0.74, pval: 1.5E-03 ! & SyRBo: 42.36s, SR: 21.08s \\
621\_fri\_c0\_100\_10 & SyRBo: 0.44, SR: 0.47, pval: 2.9E-02 ! & SyRBo: 43.37s, SR: 21.4s \\
622\_fri\_c2\_1000\_50 & SyRBo: 0.72, SR: 0.73, pval: 1.4E-01 & SyRBo: 44.87s, SR: 22.63s \\
623\_fri\_c4\_1000\_10 & SyRBo: 0.64, SR: 0.69, pval: 0.0E+00 ! & SyRBo: 53.96s, SR: 26.17s \\
624\_fri\_c0\_100\_5 & SR: 0.46, SyRBo: 0.46, pval: 9.3E-01 = & SyRBo: 43.75s, SR: 21.15s \\
626\_fri\_c2\_500\_50 & SyRBo: 0.73, SR: 0.73, pval: 2.6E-01 & SyRBo: 42.0s, SR: 21.23s \\
627\_fri\_c2\_500\_10 & SyRBo: 0.63, SR: 0.69, pval: 0.0E+00 ! & SyRBo: 52.34s, SR: 25.3s \\
628\_fri\_c3\_1000\_5 & SyRBo: 0.59, SR: 0.66, pval: 0.0E+00 ! & SyRBo: 56.7s, SR: 27.28s \\
631\_fri\_c1\_500\_5 & SyRBo: 0.6, SR: 0.68, pval: 0.0E+00 ! & SyRBo: 55.86s, SR: 26.53s \\
633\_fri\_c0\_500\_25 & SyRBo: 0.4, SR: 0.42, pval: 0.0E+00 ! & SyRBo: 51.78s, SR: 26.37s \\
634\_fri\_c2\_100\_10 & SyRBo: 0.68, SR: 0.69, pval: 9.0E-01 & SyRBo: 54.48s, SR: 26.52s \\
635\_fri\_c0\_250\_10 & SyRBo: 0.44, SR: 0.52, pval: 0.0E+00 ! & SyRBo: 43.65s, SR: 21.72s \\
637\_fri\_c1\_500\_50 & SyRBo: 0.75, SR: 0.76, pval: 1.1E-01 & SyRBo: 44.14s, SR: 22.12s \\
641\_fri\_c1\_500\_10 & SyRBo: 0.67, SR: 0.74, pval: 0.0E+00 ! & SyRBo: 54.06s, SR: 26.39s \\
643\_fri\_c2\_500\_25 & SyRBo: 0.74, SR: 0.75, pval: 1.8E-01 & SyRBo: 42.11s, SR: 20.85s \\
644\_fri\_c4\_250\_25 & SyRBo: 0.73, SR: 0.74, pval: 4.2E-01 & SyRBo: 41.93s, SR: 20.77s \\
645\_fri\_c3\_500\_50 & SyRBo: 0.7, SR: 0.7, pval: 8.6E-01 & SyRBo: 41.84s, SR: 20.81s \\
646\_fri\_c3\_500\_10 & SyRBo: 0.64, SR: 0.68, pval: 1.0E-04 ! & SyRBo: 52.88s, SR: 25.77s \\
647\_fri\_c1\_250\_10 & SyRBo: 0.65, SR: 0.73, pval: 0.0E+00 ! & SyRBo: 54.08s, SR: 25.95s \\
648\_fri\_c1\_250\_50 & SyRBo: 0.72, SR: 0.73, pval: 3.8E-01 & SyRBo: 52.98s, SR: 26.88s \\
649\_fri\_c0\_500\_5 & SyRBo: 0.4, SR: 0.46, pval: 0.0E+00 ! & SyRBo: 52.84s, SR: 25.93s \\
650\_fri\_c0\_500\_50 & SyRBo: 0.38, SR: 0.39, pval: 2.7E-02 ! & SyRBo: 52.92s, SR: 27.49s \\
651\_fri\_c0\_100\_25 & SyRBo: 0.52, SR: 0.53, pval: 4.9E-01 & SyRBo: 51.51s, SR: 25.99s \\
653\_fri\_c0\_250\_25 & SyRBo: 0.4, SR: 0.41, pval: 8.7E-03 ! & SyRBo: 53.46s, SR: 27.13s \\
654\_fri\_c0\_500\_10 & SyRBo: 0.42, SR: 0.46, pval: 0.0E+00 ! & SyRBo: 52.64s, SR: 26.29s \\
656\_fri\_c1\_100\_5 & SyRBo: 0.57, SR: 0.66, pval: 0.0E+00 ! & SyRBo: 55.37s, SR: 27.63s \\
657\_fri\_c2\_250\_10 & SyRBo: 0.63, SR: 0.7, pval: 0.0E+00 ! & SyRBo: 53.59s, SR: 25.96s \\
658\_fri\_c3\_250\_25 & SyRBo: 0.73, SR: 0.75, pval: 1.6E-01 & SyRBo: 51.63s, SR: 25.6s \\
659\_sleuth\_ex1714 & SyRBo: 6745.19, SR: 8208.12, pval: 2.2E-02 ! & SyRBo: 366.9s, SR: 179.81s \\
663\_rabe\_266 & SR: 19.76, SyRBo: 20.19, pval: 4.7E-01 = & SyRBo: 95.03s, SR: 54.36s \\
665\_sleuth\_case2002 & SR: 5.08, SyRBo: 5.33, pval: 6.8E-02 = & SyRBo: 53.61s, SR: 31.3s \\
666\_rmftsa\_ladata & SR: 1.64, SyRBo: 1.64, pval: 9.2E-01 = & SyRBo: 45.69s, SR: 25.33s \\
678\_visualizing\_environmental & SyRBo: 2.46, SR: 2.51, pval: 4.4E-01 & SyRBo: 50.56s, SR: 28.09s \\
687\_sleuth\_ex1605 & SR: 13.26, SyRBo: 14.33, pval: 5.4E-02 = & SyRBo: 90.32s, SR: 51.23s \\
690\_visualizing\_galaxy & SyRBo: 259.23, SR: 461.52, pval: 0.0E+00 ! & SyRBo: 263.2s, SR: 137.71s \\
695\_chatfield\_4 & SR: 17.47, SyRBo: 17.81, pval: 4.0E-01 = & SyRBo: 112.3s, SR: 57.87s \\
706\_sleuth\_case1202 & SR: 48.76, SyRBo: 52.21, pval: 8.1E-02 = & SyRBo: 120.17s, SR: 77.54s \\
712\_chscase\_geyser1 & SyRBo: 8.3, SR: 9.0, pval: 1.0E-04 ! & SyRBo: 71.33s, SR: 47.21s \\
\end{tabular}
\normalsize
\end{table}

\begin{table}[p]
\caption{3-stage SyRBo: Results of all datasets.}
\label{tab:allres3}
\centering
\tiny
\begin{tabular}{r|c|l}
  \textbf{dataset} & \textbf{mean absolute error and pval} & \textbf{run times} \\ \hline
1027\_ESL & SyRBo: 1.01, SR: 1.04, pval: 1.1E-02 ! & SyRBo: 70.24s, SR: 24.65s \\
1028\_SWD & SyRBo: 0.61, SR: 0.62, pval: 1.7E-01 & SyRBo: 70.79s, SR: 25.03s \\
1029\_LEV & SyRBo: 0.62, SR: 0.64, pval: 0.0E+00 ! & SyRBo: 69.2s, SR: 22.65s \\
1030\_ERA & SyRBo: 1.43, SR: 1.46, pval: 2.0E-02 ! & SyRBo: 72.45s, SR: 24.73s \\
1089\_USCrime & SyRBo: 25.31, SR: 27.01, pval: 1.8E-01 & SyRBo: 134.46s, SR: 57.85s \\
1096\_FacultySalaries & SR: 3.57, SyRBo: 3.6, pval: 7.5E-01 = & SyRBo: 74.28s, SR: 30.96s \\
192\_vineyard & SR: 2.42, SyRBo: 2.54, pval: 5.9E-02 = & SyRBo: 72.9s, SR: 30.64s \\
195\_auto\_price & SyRBo: 1955.32, SR: 2049.73, pval: 2.0E-01 & SyRBo: 558.45s, SR: 164.11s \\
207\_autoPrice & SyRBo: 1945.66, SR: 1968.41, pval: 8.6E-01 & SyRBo: 465.0s, SR: 133.33s \\
210\_cloud & SyRBo: 0.5, SR: 0.51, pval: 5.9E-01 & SyRBo: 68.09s, SR: 23.48s \\
228\_elusage & SyRBo: 12.68, SR: 14.45, pval: 2.9E-03 ! & SyRBo: 127.75s, SR: 64.49s \\
229\_pwLinear & SyRBo: 1.49, SR: 1.63, pval: 1.9E-03 ! & SyRBo: 73.73s, SR: 27.64s \\
230\_machine\_cpu & SyRBo: 40.74, SR: 44.23, pval: 1.2E-01 & SyRBo: 244.09s, SR: 101.61s \\
4544\_GeographicalOriginalofMusic & SyRBo: 0.49, SR: 0.49, pval: 9.7E-01 & SyRBo: 91.17s, SR: 38.4s \\
485\_analcatdata\_vehicle & SyRBo: 155.87, SR: 184.07, pval: 5.8E-03 ! & SyRBo: 363.3s, SR: 144.26s \\
505\_tecator & SR: 5.02, SyRBo: 5.35, pval: 1.8E-01 = & SyRBo: 140.1s, SR: 60.43s \\
519\_vinnie & SR: 1.26, SyRBo: 1.27, pval: 5.5E-01 = & SyRBo: 92.19s, SR: 35.46s \\
522\_pm10 & SyRBo: 0.67, SR: 0.69, pval: 3.1E-02 ! & SyRBo: 81.01s, SR: 27.42s \\
523\_analcatdata\_neavote & SyRBo: 0.49, SR: 0.5, pval: 5.4E-01 & SyRBo: 91.61s, SR: 39.44s \\
527\_analcatdata\_election2000 & SR: 42367.13, SyRBo: 42794.6, pval: 8.7E-01 = & SyRBo: 782.84s, SR: 187.09s \\
542\_pollution & SR: 179.19, SyRBo: 183.2, pval: 6.9E-01 = & SyRBo: 323.15s, SR: 137.38s \\
547\_no2 & SyRBo: 0.57, SR: 0.59, pval: 8.1E-03 ! & SyRBo: 84.6s, SR: 28.94s \\
556\_analcatdata\_apnea2 & SR: 838.3, SyRBo: 841.71, pval: 9.2E-01 = & SyRBo: 253.91s, SR: 91.68s \\
557\_analcatdata\_apnea1 & SR: 838.25, SyRBo: 871.41, pval: 4.8E-01 = & SyRBo: 209.53s, SR: 55.88s \\
560\_bodyfat & SR: 4.23, SyRBo: 4.34, pval: 2.1E-01 = & SyRBo: 104.88s, SR: 41.81s \\
561\_cpu & SyRBo: 30.41, SR: 33.93, pval: 1.8E-01 & SyRBo: 208.96s, SR: 89.89s \\
579\_fri\_c0\_250\_5 & SyRBo: 0.4, SR: 0.45, pval: 0.0E+00 ! & SyRBo: 79.88s, SR: 25.77s \\
581\_fri\_c3\_500\_25 & SyRBo: 0.7, SR: 0.72, pval: 3.1E-03 ! & SyRBo: 79.92s, SR: 26.46s \\
582\_fri\_c1\_500\_25 & SyRBo: 0.68, SR: 0.72, pval: 0.0E+00 ! & SyRBo: 77.19s, SR: 25.52s \\
583\_fri\_c1\_1000\_50 & SyRBo: 0.72, SR: 0.74, pval: 0.0E+00 ! & SyRBo: 75.06s, SR: 25.16s \\
584\_fri\_c4\_500\_25 & SyRBo: 0.68, SR: 0.71, pval: 0.0E+00 ! & SyRBo: 75.45s, SR: 24.92s \\
586\_fri\_c3\_1000\_25 & SyRBo: 0.68, SR: 0.7, pval: 0.0E+00 ! & SyRBo: 79.06s, SR: 26.17s \\
588\_fri\_c4\_1000\_100 & SyRBo: 0.72, SR: 0.73, pval: 1.7E-01 & SyRBo: 78.02s, SR: 26.38s \\
589\_fri\_c2\_1000\_25 & SyRBo: 0.68, SR: 0.71, pval: 0.0E+00 ! & SyRBo: 79.93s, SR: 26.46s \\
590\_fri\_c0\_1000\_50 & SyRBo: 0.37, SR: 0.41, pval: 0.0E+00 ! & SyRBo: 80.78s, SR: 28.43s \\
591\_fri\_c1\_100\_10 & SyRBo: 0.71, SR: 0.74, pval: 3.9E-02 ! & SyRBo: 79.19s, SR: 25.39s \\
592\_fri\_c4\_1000\_25 & SyRBo: 0.7, SR: 0.72, pval: 1.0E-04 ! & SyRBo: 81.93s, SR: 27.1s \\
593\_fri\_c1\_1000\_10 & SyRBo: 0.61, SR: 0.71, pval: 0.0E+00 ! & SyRBo: 82.54s, SR: 26.59s \\
594\_fri\_c2\_100\_5 & SyRBo: 0.64, SR: 0.71, pval: 0.0E+00 ! & SyRBo: 80.74s, SR: 26.65s \\
595\_fri\_c0\_1000\_10 & SyRBo: 0.35, SR: 0.44, pval: 0.0E+00 ! & SyRBo: 82.43s, SR: 27.34s \\
596\_fri\_c2\_250\_5 & SyRBo: 0.6, SR: 0.68, pval: 0.0E+00 ! & SyRBo: 80.88s, SR: 26.18s \\
597\_fri\_c2\_500\_5 & SyRBo: 0.58, SR: 0.68, pval: 0.0E+00 ! & SyRBo: 81.55s, SR: 26.48s \\
598\_fri\_c0\_1000\_25 & SyRBo: 0.36, SR: 0.43, pval: 0.0E+00 ! & SyRBo: 80.75s, SR: 27.68s \\
599\_fri\_c2\_1000\_5 & SyRBo: 0.56, SR: 0.66, pval: 0.0E+00 ! & SyRBo: 83.39s, SR: 26.86s \\
601\_fri\_c1\_250\_5 & SyRBo: 0.56, SR: 0.67, pval: 0.0E+00 ! & SyRBo: 81.61s, SR: 26.23s \\
602\_fri\_c3\_250\_10 & SyRBo: 0.68, SR: 0.72, pval: 0.0E+00 ! & SyRBo: 78.76s, SR: 25.62s \\
603\_fri\_c0\_250\_50 & SyRBo: 0.39, SR: 0.41, pval: 7.3E-03 ! & SyRBo: 77.87s, SR: 27.07s \\
604\_fri\_c4\_500\_10 & SyRBo: 0.66, SR: 0.71, pval: 0.0E+00 ! & SyRBo: 79.58s, SR: 26.2s \\
605\_fri\_c2\_250\_25 & SyRBo: 0.69, SR: 0.7, pval: 5.3E-01 & SyRBo: 79.46s, SR: 26.31s \\
606\_fri\_c2\_1000\_10 & SyRBo: 0.59, SR: 0.67, pval: 0.0E+00 ! & SyRBo: 82.15s, SR: 26.38s \\
607\_fri\_c4\_1000\_50 & SyRBo: 0.71, SR: 0.73, pval: 1.4E-02 ! & SyRBo: 78.32s, SR: 26.14s \\
608\_fri\_c3\_1000\_10 & SyRBo: 0.61, SR: 0.7, pval: 0.0E+00 ! & SyRBo: 82.63s, SR: 26.62s \\
609\_fri\_c0\_1000\_5 & SyRBo: 0.37, SR: 0.44, pval: 0.0E+00 ! & SyRBo: 79.77s, SR: 25.95s \\
611\_fri\_c3\_100\_5 & SyRBo: 0.61, SR: 0.67, pval: 8.0E-03 ! & SyRBo: 84.03s, SR: 28.26s \\
612\_fri\_c1\_1000\_5 & SyRBo: 0.55, SR: 0.69, pval: 0.0E+00 ! & SyRBo: 82.23s, SR: 26.03s \\
613\_fri\_c3\_250\_5 & SyRBo: 0.58, SR: 0.64, pval: 0.0E+00 ! & SyRBo: 79.23s, SR: 25.57s \\
615\_fri\_c4\_250\_10 & SyRBo: 0.66, SR: 0.7, pval: 0.0E+00 ! & SyRBo: 79.95s, SR: 26.18s \\
616\_fri\_c4\_500\_50 & SyRBo: 0.74, SR: 0.74, pval: 6.5E-01 & SyRBo: 80.63s, SR: 26.95s \\
617\_fri\_c3\_500\_5 & SyRBo: 0.57, SR: 0.65, pval: 0.0E+00 ! & SyRBo: 80.92s, SR: 26.7s \\
618\_fri\_c3\_1000\_50 & SyRBo: 0.71, SR: 0.73, pval: 4.0E-04 ! & SyRBo: 78.7s, SR: 26.25s \\
620\_fri\_c1\_1000\_25 & SyRBo: 0.7, SR: 0.73, pval: 0.0E+00 ! & SyRBo: 79.9s, SR: 26.4s \\
621\_fri\_c0\_100\_10 & SyRBo: 0.41, SR: 0.45, pval: 8.0E-03 ! & SyRBo: 78.91s, SR: 25.96s \\
622\_fri\_c2\_1000\_50 & SyRBo: 0.71, SR: 0.73, pval: 3.0E-04 ! & SyRBo: 66.25s, SR: 22.17s \\
623\_fri\_c4\_1000\_10 & SyRBo: 0.62, SR: 0.68, pval: 0.0E+00 ! & SyRBo: 81.44s, SR: 26.48s \\
624\_fri\_c0\_100\_5 & SyRBo: 0.42, SR: 0.47, pval: 0.0E+00 ! & SyRBo: 78.17s, SR: 25.11s \\
626\_fri\_c2\_500\_50 & SyRBo: 0.71, SR: 0.72, pval: 4.6E-01 & SyRBo: 78.22s, SR: 26.39s \\
627\_fri\_c2\_500\_10 & SyRBo: 0.61, SR: 0.69, pval: 0.0E+00 ! & SyRBo: 85.52s, SR: 27.54s \\
628\_fri\_c3\_1000\_5 & SyRBo: 0.59, SR: 0.66, pval: 0.0E+00 ! & SyRBo: 86.2s, SR: 27.97s \\
631\_fri\_c1\_500\_5 & SyRBo: 0.56, SR: 0.66, pval: 0.0E+00 ! & SyRBo: 82.42s, SR: 26.2s \\
633\_fri\_c0\_500\_25 & SyRBo: 0.37, SR: 0.43, pval: 0.0E+00 ! & SyRBo: 78.32s, SR: 26.68s \\
634\_fri\_c2\_100\_10 & SyRBo: 0.64, SR: 0.68, pval: 9.0E-04 ! & SyRBo: 80.8s, SR: 26.22s \\
635\_fri\_c0\_250\_10 & SyRBo: 0.39, SR: 0.52, pval: 0.0E+00 ! & SyRBo: 61.06s, SR: 20.16s \\
637\_fri\_c1\_500\_50 & SyRBo: 0.76, SR: 0.76, pval: 7.5E-01 & SyRBo: 79.6s, SR: 26.58s \\
641\_fri\_c1\_500\_10 & SyRBo: 0.62, SR: 0.73, pval: 0.0E+00 ! & SyRBo: 81.02s, SR: 26.14s \\
643\_fri\_c2\_500\_25 & SyRBo: 0.74, SR: 0.76, pval: 2.1E-02 ! & SyRBo: 78.8s, SR: 26.08s \\
644\_fri\_c4\_250\_25 & SyRBo: 0.72, SR: 0.74, pval: 2.0E-01 & SyRBo: 80.38s, SR: 26.48s \\
645\_fri\_c3\_500\_50 & SyRBo: 0.7, SR: 0.71, pval: 3.6E-01 & SyRBo: 80.26s, SR: 26.57s \\
646\_fri\_c3\_500\_10 & SyRBo: 0.63, SR: 0.69, pval: 0.0E+00 ! & SyRBo: 80.1s, SR: 26.1s \\
647\_fri\_c1\_250\_10 & SyRBo: 0.64, SR: 0.73, pval: 0.0E+00 ! & SyRBo: 80.7s, SR: 26.13s \\
648\_fri\_c1\_250\_50 & SyRBo: 0.74, SR: 0.74, pval: 8.4E-01 & SyRBo: 77.4s, SR: 26.33s \\
649\_fri\_c0\_500\_5 & SyRBo: 0.37, SR: 0.46, pval: 0.0E+00 ! & SyRBo: 76.0s, SR: 24.91s \\
650\_fri\_c0\_500\_50 & SyRBo: 0.37, SR: 0.4, pval: 0.0E+00 ! & SyRBo: 82.74s, SR: 28.93s \\
651\_fri\_c0\_100\_25 & SyRBo: 0.5, SR: 0.53, pval: 1.7E-02 ! & SyRBo: 77.58s, SR: 26.06s \\
653\_fri\_c0\_250\_25 & SyRBo: 0.37, SR: 0.41, pval: 0.0E+00 ! & SyRBo: 62.56s, SR: 21.2s \\
654\_fri\_c0\_500\_10 & SyRBo: 0.37, SR: 0.46, pval: 0.0E+00 ! & SyRBo: 63.3s, SR: 20.95s \\
656\_fri\_c1\_100\_5 & SyRBo: 0.55, SR: 0.65, pval: 0.0E+00 ! & SyRBo: 87.52s, SR: 28.95s \\
657\_fri\_c2\_250\_10 & SyRBo: 0.64, SR: 0.69, pval: 0.0E+00 ! & SyRBo: 84.08s, SR: 27.15s \\
658\_fri\_c3\_250\_25 & SyRBo: 0.73, SR: 0.74, pval: 1.4E-01 & SyRBo: 77.71s, SR: 25.61s \\
659\_sleuth\_ex1714 & SyRBo: 7231.48, SR: 7604.58, pval: 5.2E-01 & SyRBo: 660.4s, SR: 183.78s \\
663\_rabe\_266 & SR: 19.49, SyRBo: 20.01, pval: 2.4E-01 = & SyRBo: 167.69s, SR: 72.88s \\
665\_sleuth\_case2002 & SR: 5.3, SyRBo: 5.3, pval: 9.7E-01 = & SyRBo: 97.79s, SR: 38.17s \\
666\_rmftsa\_ladata & SyRBo: 1.58, SR: 1.62, pval: 3.0E-01 & SyRBo: 83.81s, SR: 32.08s \\
678\_visualizing\_environmental & SR: 2.44, SyRBo: 2.46, pval: 8.1E-01 = & SyRBo: 72.36s, SR: 28.73s \\
687\_sleuth\_ex1605 & SR: 13.04, SyRBo: 14.99, pval: 1.0E-04 & SyRBo: 130.96s, SR: 46.48s \\
690\_visualizing\_galaxy & SyRBo: 212.82, SR: 440.08, pval: 1.0E-04 ! & SyRBo: 408.21s, SR: 125.28s \\
695\_chatfield\_4 & SR: 16.79, SyRBo: 18.34, pval: 9.0E-03 & SyRBo: 155.94s, SR: 57.64s \\
706\_sleuth\_case1202 & SR: 49.57, SyRBo: 51.69, pval: 1.8E-01 = & SyRBo: 172.97s, SR: 75.44s \\
712\_chscase\_geyser1 & SyRBo: 8.2, SR: 8.88, pval: 0.0E+00 ! & SyRBo: 100.15s, SR: 47.06s \\
\end{tabular}
\normalsize
\end{table}

\begin{table}[p]
\caption{4-stage SyRBo: Results of all datasets.}
\label{tab:allres4}
\centering
\tiny
\begin{tabular}{r|c|l}
  \textbf{dataset} & \textbf{mean absolute error and pval} & \textbf{run times} \\ \hline
1027\_ESL & SyRBo: 1.0, SR: 1.04, pval: 2.0E-04 ! & SyRBo: 93.37s, SR: 24.88s \\
1028\_SWD & SyRBo: 0.61, SR: 0.62, pval: 1.3E-02 ! & SyRBo: 93.66s, SR: 24.91s \\
1029\_LEV & SyRBo: 0.62, SR: 0.65, pval: 0.0E+00 ! & SyRBo: 92.83s, SR: 22.9s \\
1030\_ERA & SyRBo: 1.42, SR: 1.45, pval: 2.0E-04 ! & SyRBo: 96.63s, SR: 24.49s \\
1089\_USCrime & SyRBo: 25.17, SR: 25.74, pval: 7.2E-01 & SyRBo: 168.69s, SR: 59.24s \\
1096\_FacultySalaries & SR: 3.51, SyRBo: 3.53, pval: 8.4E-01 = & SyRBo: 87.13s, SR: 27.14s \\
192\_vineyard & SR: 2.34, SyRBo: 2.55, pval: 4.0E-02 & SyRBo: 97.59s, SR: 31.41s \\
195\_auto\_price & SyRBo: 1881.2, SR: 2047.73, pval: 8.4E-03 ! & SyRBo: 606.2s, SR: 156.64s \\
207\_autoPrice & SyRBo: 1883.63, SR: 2046.39, pval: 6.0E-02 & SyRBo: 759.26s, SR: 155.12s \\
210\_cloud & SyRBo: 0.49, SR: 0.5, pval: 9.4E-01 & SyRBo: 89.57s, SR: 23.4s \\
228\_elusage & SyRBo: 12.45, SR: 14.36, pval: 6.0E-04 ! & SyRBo: 176.15s, SR: 77.55s \\
229\_pwLinear & SyRBo: 1.49, SR: 1.59, pval: 1.4E-02 ! & SyRBo: 111.62s, SR: 33.75s \\
230\_machine\_cpu & SyRBo: 43.28, SR: 47.09, pval: 7.9E-02 & SyRBo: 273.04s, SR: 91.55s \\
4544\_GeographicalOriginalofMusic & SyRBo: 0.49, SR: 0.49, pval: 2.9E-01 & SyRBo: 117.26s, SR: 38.11s \\
485\_analcatdata\_vehicle & SyRBo: 151.23, SR: 186.52, pval: 8.0E-04 ! & SyRBo: 391.71s, SR: 131.13s \\
505\_tecator & SR: 5.01, SyRBo: 5.05, pval: 9.0E-01 = & SyRBo: 161.8s, SR: 58.38s \\
519\_vinnie & SR: 1.26, SyRBo: 1.3, pval: 9.1E-02 = & SyRBo: 118.67s, SR: 35.63s \\
522\_pm10 & SyRBo: 0.66, SR: 0.69, pval: 2.0E-04 ! & SyRBo: 111.11s, SR: 28.72s \\
523\_analcatdata\_neavote & SyRBo: 0.5, SR: 0.5, pval: 9.4E-01 & SyRBo: 114.11s, SR: 37.29s \\
527\_analcatdata\_election2000 & SR: 41409.25, SyRBo: 43867.25, pval: 4.9E-01 = & SyRBo: 865.92s, SR: 160.52s \\
542\_pollution & SyRBo: 180.88, SR: 188.26, pval: 4.4E-01 & SyRBo: 367.71s, SR: 141.35s \\
547\_no2 & SyRBo: 0.56, SR: 0.59, pval: 4.0E-04 ! & SyRBo: 109.17s, SR: 28.85s \\
556\_analcatdata\_apnea2 & SR: 869.07, SyRBo: 881.56, pval: 8.6E-01 = & SyRBo: 238.22s, SR: 75.84s \\
557\_analcatdata\_apnea1 & SyRBo: 861.47, SR: 869.01, pval: 9.1E-01 & SyRBo: 215.12s, SR: 54.48s \\
560\_bodyfat & SR: 4.24, SyRBo: 4.37, pval: 3.0E-01 = & SyRBo: 129.35s, SR: 40.74s \\
561\_cpu & SyRBo: 29.33, SR: 35.67, pval: 3.6E-03 ! & SyRBo: 254.82s, SR: 95.29s \\
579\_fri\_c0\_250\_5 & SyRBo: 0.38, SR: 0.45, pval: 0.0E+00 ! & SyRBo: 83.29s, SR: 20.24s \\
581\_fri\_c3\_500\_25 & SyRBo: 0.68, SR: 0.72, pval: 0.0E+00 ! & SyRBo: 107.74s, SR: 26.62s \\
582\_fri\_c1\_500\_25 & SyRBo: 0.66, SR: 0.72, pval: 0.0E+00 ! & SyRBo: 83.86s, SR: 20.72s \\
583\_fri\_c1\_1000\_50 & SyRBo: 0.7, SR: 0.74, pval: 0.0E+00 ! & SyRBo: 107.04s, SR: 26.75s \\
584\_fri\_c4\_500\_25 & SyRBo: 0.67, SR: 0.72, pval: 0.0E+00 ! & SyRBo: 106.2s, SR: 26.11s \\
586\_fri\_c3\_1000\_25 & SyRBo: 0.66, SR: 0.71, pval: 0.0E+00 ! & SyRBo: 106.28s, SR: 26.3s \\
588\_fri\_c4\_1000\_100 & SyRBo: 0.72, SR: 0.72, pval: 6.5E-01 & SyRBo: 105.29s, SR: 26.7s \\
589\_fri\_c2\_1000\_25 & SyRBo: 0.67, SR: 0.71, pval: 0.0E+00 ! & SyRBo: 89.31s, SR: 22.21s \\
590\_fri\_c0\_1000\_50 & SyRBo: 0.36, SR: 0.4, pval: 0.0E+00 ! & SyRBo: 107.96s, SR: 28.67s \\
591\_fri\_c1\_100\_10 & SyRBo: 0.68, SR: 0.74, pval: 6.6E-03 ! & SyRBo: 107.71s, SR: 26.22s \\
592\_fri\_c4\_1000\_25 & SyRBo: 0.69, SR: 0.72, pval: 0.0E+00 ! & SyRBo: 104.6s, SR: 25.92s \\
593\_fri\_c1\_1000\_10 & SyRBo: 0.58, SR: 0.71, pval: 0.0E+00 ! & SyRBo: 84.35s, SR: 20.31s \\
594\_fri\_c2\_100\_5 & SyRBo: 0.62, SR: 0.7, pval: 0.0E+00 ! & SyRBo: 86.38s, SR: 21.38s \\
595\_fri\_c0\_1000\_10 & SyRBo: 0.33, SR: 0.44, pval: 0.0E+00 ! & SyRBo: 110.51s, SR: 27.53s \\
596\_fri\_c2\_250\_5 & SyRBo: 0.59, SR: 0.69, pval: 0.0E+00 ! & SyRBo: 106.65s, SR: 26.23s \\
597\_fri\_c2\_500\_5 & SyRBo: 0.57, SR: 0.67, pval: 0.0E+00 ! & SyRBo: 106.16s, SR: 26.2s \\
598\_fri\_c0\_1000\_25 & SyRBo: 0.35, SR: 0.43, pval: 0.0E+00 ! & SyRBo: 106.81s, SR: 27.65s \\
599\_fri\_c2\_1000\_5 & SyRBo: 0.54, SR: 0.67, pval: 0.0E+00 ! & SyRBo: 110.59s, SR: 26.69s \\
601\_fri\_c1\_250\_5 & SyRBo: 0.53, SR: 0.66, pval: 0.0E+00 ! & SyRBo: 106.99s, SR: 26.05s \\
602\_fri\_c3\_250\_10 & SyRBo: 0.66, SR: 0.73, pval: 0.0E+00 ! & SyRBo: 107.64s, SR: 26.21s \\
603\_fri\_c0\_250\_50 & SyRBo: 0.39, SR: 0.4, pval: 2.9E-02 ! & SyRBo: 104.74s, SR: 27.57s \\
604\_fri\_c4\_500\_10 & SyRBo: 0.63, SR: 0.71, pval: 0.0E+00 ! & SyRBo: 105.4s, SR: 25.79s \\
605\_fri\_c2\_250\_25 & SyRBo: 0.68, SR: 0.7, pval: 6.7E-02 & SyRBo: 103.7s, SR: 25.76s \\
606\_fri\_c2\_1000\_10 & SyRBo: 0.57, SR: 0.68, pval: 0.0E+00 ! & SyRBo: 110.65s, SR: 26.65s \\
607\_fri\_c4\_1000\_50 & SyRBo: 0.72, SR: 0.73, pval: 6.9E-02 & SyRBo: 111.03s, SR: 27.78s \\
608\_fri\_c3\_1000\_10 & SyRBo: 0.59, SR: 0.7, pval: 0.0E+00 ! & SyRBo: 118.4s, SR: 28.63s \\
609\_fri\_c0\_1000\_5 & SyRBo: 0.34, SR: 0.44, pval: 0.0E+00 ! & SyRBo: 116.25s, SR: 28.29s \\
611\_fri\_c3\_100\_5 & SyRBo: 0.62, SR: 0.65, pval: 1.3E-02 ! & SyRBo: 106.5s, SR: 26.92s \\
612\_fri\_c1\_1000\_5 & SyRBo: 0.54, SR: 0.69, pval: 0.0E+00 ! & SyRBo: 95.28s, SR: 22.96s \\
613\_fri\_c3\_250\_5 & SyRBo: 0.56, SR: 0.64, pval: 0.0E+00 ! & SyRBo: 85.81s, SR: 20.96s \\
615\_fri\_c4\_250\_10 & SyRBo: 0.64, SR: 0.7, pval: 0.0E+00 ! & SyRBo: 103.59s, SR: 25.5s \\
616\_fri\_c4\_500\_50 & SyRBo: 0.73, SR: 0.74, pval: 7.3E-01 & SyRBo: 101.64s, SR: 25.4s \\
617\_fri\_c3\_500\_5 & SyRBo: 0.55, SR: 0.66, pval: 0.0E+00 ! & SyRBo: 106.97s, SR: 26.56s \\
618\_fri\_c3\_1000\_50 & SyRBo: 0.7, SR: 0.72, pval: 0.0E+00 ! & SyRBo: 106.08s, SR: 26.47s \\
620\_fri\_c1\_1000\_25 & SyRBo: 0.68, SR: 0.74, pval: 0.0E+00 ! & SyRBo: 109.58s, SR: 27.21s \\
621\_fri\_c0\_100\_10 & SyRBo: 0.4, SR: 0.45, pval: 0.0E+00 ! & SyRBo: 103.48s, SR: 25.55s \\
622\_fri\_c2\_1000\_50 & SyRBo: 0.71, SR: 0.73, pval: 6.0E-04 ! & SyRBo: 106.03s, SR: 26.9s \\
623\_fri\_c4\_1000\_10 & SyRBo: 0.6, SR: 0.69, pval: 0.0E+00 ! & SyRBo: 109.83s, SR: 26.66s \\
624\_fri\_c0\_100\_5 & SyRBo: 0.41, SR: 0.47, pval: 0.0E+00 ! & SyRBo: 110.93s, SR: 26.78s \\
626\_fri\_c2\_500\_50 & SyRBo: 0.72, SR: 0.73, pval: 8.7E-02 & SyRBo: 104.49s, SR: 26.41s \\
627\_fri\_c2\_500\_10 & SyRBo: 0.58, SR: 0.69, pval: 0.0E+00 ! & SyRBo: 106.73s, SR: 25.88s \\
628\_fri\_c3\_1000\_5 & SyRBo: 0.58, SR: 0.66, pval: 0.0E+00 ! & SyRBo: 108.33s, SR: 26.76s \\
631\_fri\_c1\_500\_5 & SyRBo: 0.54, SR: 0.67, pval: 0.0E+00 ! & SyRBo: 107.88s, SR: 26.04s \\
633\_fri\_c0\_500\_25 & SyRBo: 0.34, SR: 0.42, pval: 0.0E+00 ! & SyRBo: 104.15s, SR: 26.69s \\
634\_fri\_c2\_100\_10 & SyRBo: 0.67, SR: 0.71, pval: 7.3E-03 ! & SyRBo: 106.29s, SR: 26.07s \\
635\_fri\_c0\_250\_10 & SyRBo: 0.37, SR: 0.52, pval: 0.0E+00 ! & SyRBo: 102.86s, SR: 25.5s \\
637\_fri\_c1\_500\_50 & SyRBo: 0.74, SR: 0.76, pval: 5.0E-04 ! & SyRBo: 104.06s, SR: 26.12s \\
641\_fri\_c1\_500\_10 & SyRBo: 0.59, SR: 0.74, pval: 0.0E+00 ! & SyRBo: 108.72s, SR: 26.43s \\
643\_fri\_c2\_500\_25 & SyRBo: 0.71, SR: 0.76, pval: 0.0E+00 ! & SyRBo: 105.35s, SR: 26.16s \\
644\_fri\_c4\_250\_25 & SyRBo: 0.72, SR: 0.75, pval: 4.9E-03 ! & SyRBo: 104.33s, SR: 25.79s \\
645\_fri\_c3\_500\_50 & SR: 0.7, SyRBo: 0.7, pval: 7.6E-01 = & SyRBo: 103.7s, SR: 25.8s \\
646\_fri\_c3\_500\_10 & SyRBo: 0.61, SR: 0.69, pval: 0.0E+00 ! & SyRBo: 91.22s, SR: 22.33s \\
647\_fri\_c1\_250\_10 & SyRBo: 0.61, SR: 0.74, pval: 0.0E+00 ! & SyRBo: 103.42s, SR: 25.12s \\
648\_fri\_c1\_250\_50 & SyRBo: 0.72, SR: 0.75, pval: 2.1E-03 ! & SyRBo: 108.59s, SR: 27.71s \\
649\_fri\_c0\_500\_5 & SyRBo: 0.34, SR: 0.46, pval: 0.0E+00 ! & SyRBo: 112.89s, SR: 27.46s \\
650\_fri\_c0\_500\_50 & SyRBo: 0.36, SR: 0.39, pval: 0.0E+00 ! & SyRBo: 85.82s, SR: 22.59s \\
651\_fri\_c0\_100\_25 & SyRBo: 0.51, SR: 0.53, pval: 4.4E-02 ! & SyRBo: 111.2s, SR: 28.12s \\
653\_fri\_c0\_250\_25 & SyRBo: 0.36, SR: 0.41, pval: 0.0E+00 ! & SyRBo: 83.61s, SR: 21.3s \\
654\_fri\_c0\_500\_10 & SyRBo: 0.35, SR: 0.46, pval: 0.0E+00 ! & SyRBo: 84.78s, SR: 21.04s \\
656\_fri\_c1\_100\_5 & SyRBo: 0.54, SR: 0.64, pval: 0.0E+00 ! & SyRBo: 86.37s, SR: 21.66s \\
657\_fri\_c2\_250\_10 & SyRBo: 0.63, SR: 0.69, pval: 0.0E+00 ! & SyRBo: 105.03s, SR: 25.64s \\
658\_fri\_c3\_250\_25 & SyRBo: 0.72, SR: 0.75, pval: 8.2E-03 ! & SyRBo: 103.48s, SR: 25.53s \\
659\_sleuth\_ex1714 & SyRBo: 6437.7, SR: 7641.02, pval: 7.0E-03 ! & SyRBo: 723.0s, SR: 171.93s \\
663\_rabe\_266 & SR: 19.85, SyRBo: 20.79, pval: 6.7E-02 = & SyRBo: 188.09s, SR: 66.1s \\
665\_sleuth\_case2002 & SR: 5.06, SyRBo: 5.15, pval: 4.3E-01 = & SyRBo: 125.81s, SR: 37.11s \\
666\_rmftsa\_ladata & SyRBo: 1.58, SR: 1.65, pval: 3.1E-02 ! & SyRBo: 110.41s, SR: 31.81s \\
678\_visualizing\_environmental & SR: 2.44, SyRBo: 2.46, pval: 8.1E-01 = & SyRBo: 116.57s, SR: 35.84s \\
687\_sleuth\_ex1605 & SR: 13.27, SyRBo: 15.03, pval: 1.5E-03 & SyRBo: 187.9s, SR: 58.29s \\
690\_visualizing\_galaxy & SyRBo: 219.9, SR: 470.38, pval: 0.0E+00 ! & SyRBo: 599.04s, SR: 171.58s \\
695\_chatfield\_4 & SR: 16.84, SyRBo: 17.27, pval: 4.3E-01 = & SyRBo: 245.5s, SR: 82.54s \\
706\_sleuth\_case1202 & SR: 46.85, SyRBo: 48.25, pval: 2.8E-01 = & SyRBo: 286.46s, SR: 95.73s \\
712\_chscase\_geyser1 & SyRBo: 8.27, SR: 9.11, pval: 0.0E+00 ! & SyRBo: 114.63s, SR: 43.05s \\
\end{tabular}
\normalsize
\end{table}

\begin{table}[p]
\caption{5-stage SyRBo: Results of all datasets.}
\label{tab:allres5}
\centering
\tiny
\begin{tabular}{r|c|l}
  \textbf{dataset} & \textbf{mean absolute error and pval} & \textbf{run times} \\ \hline
1027\_ESL & SyRBo: 0.99, SR: 1.04, pval: 0.0E+00 ! & SyRBo: 138.45s, SR: 30.02s \\
1028\_SWD & SyRBo: 0.6, SR: 0.62, pval: 3.2E-02 ! & SyRBo: 112.18s, SR: 24.21s \\
1029\_LEV & SyRBo: 0.61, SR: 0.65, pval: 0.0E+00 ! & SyRBo: 108.52s, SR: 21.54s \\
1030\_ERA & SyRBo: 1.41, SR: 1.46, pval: 0.0E+00 ! & SyRBo: 114.87s, SR: 23.88s \\
1089\_USCrime & SR: 25.79, SyRBo: 26.52, pval: 6.2E-01 = & SyRBo: 244.05s, SR: 70.49s \\
1096\_FacultySalaries & SR: 3.44, SyRBo: 3.66, pval: 2.0E-01 = & SyRBo: 137.41s, SR: 36.12s \\
192\_vineyard & SR: 2.45, SyRBo: 2.53, pval: 2.7E-01 = & SyRBo: 145.21s, SR: 37.57s \\
195\_auto\_price & SyRBo: 1846.49, SR: 1984.64, pval: 2.5E-02 ! & SyRBo: 1136.73s, SR: 181.96s \\
207\_autoPrice & SyRBo: 1917.2, SR: 2004.04, pval: 3.0E-01 & SyRBo: 840.85s, SR: 146.49s \\
210\_cloud & SyRBo: 0.5, SR: 0.51, pval: 5.9E-01 & SyRBo: 136.83s, SR: 28.6s \\
228\_elusage & SyRBo: 11.83, SR: 13.25, pval: 8.6E-03 ! & SyRBo: 168.85s, SR: 59.4s \\
229\_pwLinear & SyRBo: 1.49, SR: 1.58, pval: 4.6E-02 ! & SyRBo: 147.58s, SR: 35.66s \\
230\_machine\_cpu & SyRBo: 41.55, SR: 47.64, pval: 6.6E-03 ! & SyRBo: 302.39s, SR: 89.29s \\
4544\_GeographicalOriginalofMusic & SyRBo: 0.49, SR: 0.5, pval: 5.6E-01 & SyRBo: 139.81s, SR: 37.08s \\
485\_analcatdata\_vehicle & SyRBo: 152.36, SR: 185.49, pval: 6.0E-04 ! & SyRBo: 402.2s, SR: 123.11s \\
505\_tecator & SyRBo: 5.03, SR: 5.42, pval: 9.3E-02 & SyRBo: 192.01s, SR: 56.36s \\
519\_vinnie & SR: 1.26, SyRBo: 1.28, pval: 2.0E-01 = & SyRBo: 144.14s, SR: 34.81s \\
522\_pm10 & SyRBo: 0.66, SR: 0.69, pval: 0.0E+00 ! & SyRBo: 109.71s, SR: 22.76s \\
523\_analcatdata\_neavote & SyRBo: 0.5, SR: 0.51, pval: 2.4E-01 & SyRBo: 145.96s, SR: 40.78s \\
527\_analcatdata\_election2000 & SyRBo: 41978.87, SR: 42335.38, pval: 9.2E-01 & SyRBo: 1276.47s, SR: 189.6s \\
542\_pollution & SyRBo: 184.53, SR: 186.47, pval: 8.6E-01 & SyRBo: 482.44s, SR: 152.25s \\
547\_no2 & SyRBo: 0.56, SR: 0.59, pval: 1.2E-03 ! & SyRBo: 149.77s, SR: 32.23s \\
556\_analcatdata\_apnea2 & SyRBo: 831.42, SR: 837.88, pval: 9.0E-01 & SyRBo: 374.25s, SR: 108.94s \\
557\_analcatdata\_apnea1 & SR: 833.02, SyRBo: 845.73, pval: 7.5E-01 = & SyRBo: 333.67s, SR: 66.34s \\
560\_bodyfat & SyRBo: 4.26, SR: 4.28, pval: 9.1E-01 & SyRBo: 127.64s, SR: 31.6s \\
561\_cpu & SyRBo: 28.02, SR: 32.76, pval: 2.5E-02 ! & SyRBo: 268.56s, SR: 94.84s \\
579\_fri\_c0\_250\_5 & SyRBo: 0.38, SR: 0.45, pval: 0.0E+00 ! & SyRBo: 127.54s, SR: 24.79s \\
581\_fri\_c3\_500\_25 & SyRBo: 0.68, SR: 0.72, pval: 0.0E+00 ! & SyRBo: 105.16s, SR: 20.78s \\
582\_fri\_c1\_500\_25 & SyRBo: 0.66, SR: 0.71, pval: 0.0E+00 ! & SyRBo: 104.49s, SR: 20.64s \\
583\_fri\_c1\_1000\_50 & SyRBo: 0.7, SR: 0.74, pval: 0.0E+00 ! & SyRBo: 109.61s, SR: 21.76s \\
584\_fri\_c4\_500\_25 & SyRBo: 0.66, SR: 0.71, pval: 0.0E+00 ! & SyRBo: 131.15s, SR: 26.05s \\
586\_fri\_c3\_1000\_25 & SyRBo: 0.67, SR: 0.7, pval: 0.0E+00 ! & SyRBo: 111.2s, SR: 21.99s \\
588\_fri\_c4\_1000\_100 & SyRBo: 0.72, SR: 0.72, pval: 5.1E-01 & SyRBo: 109.24s, SR: 21.66s \\
589\_fri\_c2\_1000\_25 & SyRBo: 0.67, SR: 0.71, pval: 0.0E+00 ! & SyRBo: 131.08s, SR: 26.12s \\
590\_fri\_c0\_1000\_50 & SyRBo: 0.35, SR: 0.4, pval: 0.0E+00 ! & SyRBo: 130.53s, SR: 27.88s \\
591\_fri\_c1\_100\_10 & SyRBo: 0.68, SR: 0.74, pval: 1.1E-02 ! & SyRBo: 108.96s, SR: 21.35s \\
592\_fri\_c4\_1000\_25 & SyRBo: 0.68, SR: 0.72, pval: 0.0E+00 ! & SyRBo: 106.21s, SR: 20.99s \\
593\_fri\_c1\_1000\_10 & SyRBo: 0.55, SR: 0.71, pval: 0.0E+00 ! & SyRBo: 108.31s, SR: 20.94s \\
594\_fri\_c2\_100\_5 & SyRBo: 0.61, SR: 0.68, pval: 0.0E+00 ! & SyRBo: 111.13s, SR: 22.03s \\
595\_fri\_c0\_1000\_10 & SyRBo: 0.31, SR: 0.44, pval: 0.0E+00 ! & SyRBo: 106.28s, SR: 21.17s \\
596\_fri\_c2\_250\_5 & SyRBo: 0.58, SR: 0.69, pval: 0.0E+00 ! & SyRBo: 104.44s, SR: 20.42s \\
597\_fri\_c2\_500\_5 & SyRBo: 0.55, SR: 0.67, pval: 0.0E+00 ! & SyRBo: 106.51s, SR: 20.93s \\
598\_fri\_c0\_1000\_25 & SyRBo: 0.33, SR: 0.43, pval: 0.0E+00 ! & SyRBo: 107.87s, SR: 22.01s \\
599\_fri\_c2\_1000\_5 & SyRBo: 0.53, SR: 0.67, pval: 0.0E+00 ! & SyRBo: 107.85s, SR: 20.91s \\
601\_fri\_c1\_250\_5 & SyRBo: 0.52, SR: 0.66, pval: 0.0E+00 ! & SyRBo: 132.58s, SR: 25.88s \\
602\_fri\_c3\_250\_10 & SyRBo: 0.63, SR: 0.72, pval: 0.0E+00 ! & SyRBo: 103.12s, SR: 19.87s \\
603\_fri\_c0\_250\_50 & SyRBo: 0.38, SR: 0.41, pval: 0.0E+00 ! & SyRBo: 104.63s, SR: 21.89s \\
604\_fri\_c4\_500\_10 & SyRBo: 0.62, SR: 0.72, pval: 0.0E+00 ! & SyRBo: 107.77s, SR: 21.26s \\
605\_fri\_c2\_250\_25 & SyRBo: 0.67, SR: 0.7, pval: 5.1E-03 ! & SyRBo: 100.7s, SR: 20.1s \\
606\_fri\_c2\_1000\_10 & SyRBo: 0.56, SR: 0.68, pval: 0.0E+00 ! & SyRBo: 132.9s, SR: 25.78s \\
607\_fri\_c4\_1000\_50 & SyRBo: 0.71, SR: 0.73, pval: 3.3E-02 ! & SyRBo: 128.27s, SR: 25.68s \\
608\_fri\_c3\_1000\_10 & SyRBo: 0.57, SR: 0.7, pval: 0.0E+00 ! & SyRBo: 137.65s, SR: 26.7s \\
609\_fri\_c0\_1000\_5 & SyRBo: 0.33, SR: 0.44, pval: 0.0E+00 ! & SyRBo: 137.83s, SR: 26.65s \\
611\_fri\_c3\_100\_5 & SyRBo: 0.61, SR: 0.66, pval: 3.3E-03 ! & SyRBo: 133.97s, SR: 27.56s \\
612\_fri\_c1\_1000\_5 & SyRBo: 0.53, SR: 0.68, pval: 0.0E+00 ! & SyRBo: 142.24s, SR: 27.26s \\
613\_fri\_c3\_250\_5 & SyRBo: 0.56, SR: 0.65, pval: 0.0E+00 ! & SyRBo: 138.42s, SR: 27.06s \\
615\_fri\_c4\_250\_10 & SyRBo: 0.64, SR: 0.7, pval: 0.0E+00 ! & SyRBo: 105.66s, SR: 20.78s \\
616\_fri\_c4\_500\_50 & SyRBo: 0.73, SR: 0.74, pval: 3.6E-01 & SyRBo: 129.48s, SR: 25.82s \\
617\_fri\_c3\_500\_5 & SyRBo: 0.55, SR: 0.64, pval: 0.0E+00 ! & SyRBo: 134.39s, SR: 26.93s \\
618\_fri\_c3\_1000\_50 & SyRBo: 0.71, SR: 0.73, pval: 1.8E-03 ! & SyRBo: 128.38s, SR: 25.71s \\
620\_fri\_c1\_1000\_25 & SyRBo: 0.66, SR: 0.74, pval: 0.0E+00 ! & SyRBo: 128.35s, SR: 25.35s \\
621\_fri\_c0\_100\_10 & SyRBo: 0.39, SR: 0.47, pval: 0.0E+00 ! & SyRBo: 125.76s, SR: 24.98s \\
622\_fri\_c2\_1000\_50 & SyRBo: 0.7, SR: 0.73, pval: 0.0E+00 ! & SyRBo: 130.39s, SR: 26.39s \\
623\_fri\_c4\_1000\_10 & SyRBo: 0.58, SR: 0.69, pval: 0.0E+00 ! & SyRBo: 137.99s, SR: 26.69s \\
624\_fri\_c0\_100\_5 & SyRBo: 0.41, SR: 0.47, pval: 0.0E+00 ! & SyRBo: 132.62s, SR: 25.69s \\
626\_fri\_c2\_500\_50 & SyRBo: 0.71, SR: 0.73, pval: 8.3E-02 & SyRBo: 127.3s, SR: 25.7s \\
627\_fri\_c2\_500\_10 & SyRBo: 0.58, SR: 0.69, pval: 0.0E+00 ! & SyRBo: 129.13s, SR: 25.24s \\
628\_fri\_c3\_1000\_5 & SyRBo: 0.57, SR: 0.66, pval: 0.0E+00 ! & SyRBo: 108.94s, SR: 21.41s \\
631\_fri\_c1\_500\_5 & SyRBo: 0.54, SR: 0.68, pval: 0.0E+00 ! & SyRBo: 136.54s, SR: 26.34s \\
633\_fri\_c0\_500\_25 & SyRBo: 0.33, SR: 0.43, pval: 0.0E+00 ! & SyRBo: 101.34s, SR: 20.79s \\
634\_fri\_c2\_100\_10 & SyRBo: 0.64, SR: 0.7, pval: 8.9E-03 ! & SyRBo: 135.04s, SR: 26.73s \\
635\_fri\_c0\_250\_10 & SyRBo: 0.36, SR: 0.51, pval: 0.0E+00 ! & SyRBo: 132.75s, SR: 26.19s \\
637\_fri\_c1\_500\_50 & SyRBo: 0.75, SR: 0.76, pval: 1.6E-01 & SyRBo: 132.3s, SR: 26.57s \\
641\_fri\_c1\_500\_10 & SyRBo: 0.57, SR: 0.73, pval: 0.0E+00 ! & SyRBo: 128.61s, SR: 25.01s \\
643\_fri\_c2\_500\_25 & SyRBo: 0.72, SR: 0.75, pval: 8.3E-03 ! & SyRBo: 105.24s, SR: 20.85s \\
644\_fri\_c4\_250\_25 & SyRBo: 0.7, SR: 0.74, pval: 2.5E-03 ! & SyRBo: 129.54s, SR: 25.53s \\
645\_fri\_c3\_500\_50 & SyRBo: 0.69, SR: 0.71, pval: 2.4E-02 ! & SyRBo: 125.64s, SR: 24.97s \\
646\_fri\_c3\_500\_10 & SyRBo: 0.58, SR: 0.69, pval: 0.0E+00 ! & SyRBo: 134.8s, SR: 26.19s \\
647\_fri\_c1\_250\_10 & SyRBo: 0.6, SR: 0.74, pval: 0.0E+00 ! & SyRBo: 105.11s, SR: 20.46s \\
648\_fri\_c1\_250\_50 & SyRBo: 0.73, SR: 0.74, pval: 5.4E-01 & SyRBo: 128.78s, SR: 26.5s \\
649\_fri\_c0\_500\_5 & SyRBo: 0.34, SR: 0.46, pval: 0.0E+00 ! & SyRBo: 129.74s, SR: 25.21s \\
650\_fri\_c0\_500\_50 & SyRBo: 0.34, SR: 0.39, pval: 0.0E+00 ! & SyRBo: 128.54s, SR: 27.19s \\
651\_fri\_c0\_100\_25 & SyRBo: 0.51, SR: 0.53, pval: 4.6E-02 ! & SyRBo: 131.91s, SR: 26.73s \\
653\_fri\_c0\_250\_25 & SyRBo: 0.35, SR: 0.41, pval: 0.0E+00 ! & SyRBo: 127.19s, SR: 26.07s \\
654\_fri\_c0\_500\_10 & SyRBo: 0.34, SR: 0.46, pval: 0.0E+00 ! & SyRBo: 108.92s, SR: 21.67s \\
656\_fri\_c1\_100\_5 & SyRBo: 0.55, SR: 0.64, pval: 0.0E+00 ! & SyRBo: 141.71s, SR: 28.99s \\
657\_fri\_c2\_250\_10 & SyRBo: 0.62, SR: 0.68, pval: 0.0E+00 ! & SyRBo: 129.88s, SR: 25.33s \\
658\_fri\_c3\_250\_25 & SyRBo: 0.73, SR: 0.74, pval: 1.2E-01 & SyRBo: 125.15s, SR: 24.64s \\
659\_sleuth\_ex1714 & SyRBo: 6464.12, SR: 7876.73, pval: 5.2E-02 & SyRBo: 976.99s, SR: 183.76s \\
663\_rabe\_266 & SR: 19.33, SyRBo: 20.57, pval: 1.4E-02 & SyRBo: 212.59s, SR: 60.57s \\
665\_sleuth\_case2002 & SR: 5.18, SyRBo: 5.37, pval: 4.0E-01 = & SyRBo: 148.57s, SR: 37.37s \\
666\_rmftsa\_ladata & SyRBo: 1.59, SR: 1.63, pval: 1.6E-01 & SyRBo: 134.26s, SR: 32.33s \\
678\_visualizing\_environmental & SR: 2.42, SyRBo: 2.49, pval: 6.4E-01 = & SyRBo: 138.89s, SR: 34.34s \\
687\_sleuth\_ex1605 & SR: 13.24, SyRBo: 14.84, pval: 1.1E-03 & SyRBo: 217.53s, SR: 61.9s \\
690\_visualizing\_galaxy & SyRBo: 205.0, SR: 444.22, pval: 0.0E+00 ! & SyRBo: 739.6s, SR: 171.47s \\
695\_chatfield\_4 & SR: 17.49, SyRBo: 18.39, pval: 6.1E-02 = & SyRBo: 253.22s, SR: 80.88s \\
706\_sleuth\_case1202 & SR: 48.42, SyRBo: 52.14, pval: 1.5E-01 = & SyRBo: 331.85s, SR: 85.21s \\
712\_chscase\_geyser1 & SyRBo: 8.32, SR: 8.9, pval: 3.0E-04 ! & SyRBo: 178.29s, SR: 57.9s \\
\end{tabular}
\normalsize
\end{table}

\end{document}